\pdfoutput=1

\documentclass[11pt]{article}

\usepackage{EMNLP2023}

\usepackage{times}
\usepackage{latexsym}

\usepackage[T1]{fontenc}

\usepackage[utf8]{inputenc}

\usepackage{microtype}

\usepackage{inconsolata}

\usepackage{algorithm}
\usepackage{algpseudocode}
\usepackage{bm}
\usepackage{bbm}
\usepackage{booktabs}
\usepackage[subrefformat=parens]{subcaption}
\usepackage{xspace}
\usepackage{graphicx}
\usepackage{url}
\usepackage{footmisc}
\usepackage[switch,mathlines]{lineno}
\usepackage{amsmath}
\usepackage{amssymb}
\usepackage{amsfonts}
\usepackage{etoolbox}

\newcommand*\linenomathpatch[1]{%
  \cspreto{#1}{\linenomath}%
  \cspreto{#1*}{\linenomath}%
  \csappto{end#1}{\endlinenomath}%
  \csappto{end#1*}{\endlinenomath}%
}
\newcommand*\linenomathpatchAMS[1]{%
  \cspreto{#1}{\linenomathAMS}%
  \cspreto{#1*}{\linenomathAMS}%
  \csappto{end#1}{\endlinenomath}%
  \csappto{end#1*}{\endlinenomath}%
}

\expandafter\ifx\linenomath\linenomathWithnumbers
  \let\linenomathAMS\linenomathWithnumbers
  \patchcmd\linenomathAMS{\advance\postdisplaypenalty\linenopenalty}{}{}{}
\else
  \let\linenomathAMS\linenomathNonumbers
\fi

\linenomathpatch{equation}
\linenomathpatchAMS{gather}
\linenomathpatchAMS{multline}
\linenomathpatchAMS{align}
\linenomathpatchAMS{alignat}
\linenomathpatchAMS{flalign}

\makeatletter
\patchcmd{\mmeasure@}{\measuring@true}{
  \measuring@true
  \ifnum-\linenopenaltypar>\interdisplaylinepenalty
    \advance\interdisplaylinepenalty-\linenopenalty
  \fi
  }{}{}
\makeatother

\DeclareMathOperator*{\argmin}{argmin}
\newcommand{\knn}{$k$NN\xspace}
\newcommand{\knnmt}{$k$NN-MT\xspace}
\newcommand{\pknn}{p_{k\mathrm{NN}}}
\newcommand{\pmt}{p_\mathrm{MT}}
\newcommand{\knearestneighbor}{$k$-nearest-neighbor\xspace}
\newcommand{\nnearestneighbor}{$n$-nearest-neighbor\xspace}
\newcommand{\knnseq}{\textsc{knn-seq}\xspace}
\newcommand{\fairseq}{\textsc{fairseq}\xspace}
\newcommand{\faiss}{\textsc{faiss}\xspace}

\title{\knnseq: Efficient, Extensible \knnmt Framework}

\author{
Hiroyuki Deguchi${}^{1, 2}$ \
Hayate Hirano${}^{1}$ \
Tomoki Hoshino${}^{3}$ \
 \\
\textbf{
Yuto Nishida${}^{1}$ \
Justin Vasselli${}^{1}$ \
Taro Watanabe${}^{1}$ \
}
\\
${}^{1}$Nara Institute of Science and Technology \
${}^{3}$HAKUHODO Technologies Inc. \\
${}^{2}$National Institute of Information and Communications Technology \\
 \
\texttt{\{deguchi.hiroyuki.db0, hirano.hayate.hc2, nishida.yuto.nu8,} \\
\texttt{vasselli.justin\_ray.vk4, taro\}@is.naist.jp} \\
\texttt{tomoki.hoshino@hakuhodo-technologies.co.jp} \
}

\begin{document}
\maketitle
\begin{abstract}
\knearestneighbor machine translation (\knnmt)~\cite{khandelwal-etal-2021-nearest} boosts the translation quality of a pre-trained neural machine translation (NMT) model by utilizing translation examples during decoding.
Translation examples are stored in a vector database, called a datastore, which contains one entry for each target token from the parallel data it is made from. 
Due to its size, it is computationally expensive both to construct and to retrieve examples from the datastore.
In this paper, we present an efficient and extensible \knnmt framework, \knnseq, for researchers and developers that is carefully designed to run efficiently, even with a billion-scale large datastore.
\knnseq is developed as a plug-in on \fairseq and easy to switch models and \knn indexes.
Experimental results show that our implemented \knnmt achieves a comparable gain to the original \knnmt, and the billion-scale datastore construction took 2.21 hours in the WMT'19 German-to-English translation task.
We publish our \knnseq as an MIT-licensed open-source project and the code is available on GitHub.\footnote{\url{https://github.com/naist-nlp/knn-seq}\label{footer:github}}
The demo video is available on YouTube.\footnote{\url{https://youtu.be/zTDzEOq80m0}\label{footer:youtube}}
\end{abstract}

\section{Introduction}
Neural machine translation (NMT) has achieved state-of-the-art translation performance and is attracting attention from both academia and industry~\cite{sutskever-etal-2014-sequence,bahdanau-etal-2015-neural,luong-etal-2015-effective,wu-etal-2016-google,vaswani-etal-2017-attention}.
Recently, \knearestneighbor machine translation (\knnmt)~\cite{khandelwal-etal-2021-nearest} has become a popular method to improve the translation quality of a pre-trained neural machine translation (NMT) model by using  translation examples during decoding.
This technique has been particularly successful in domain adaption, improving translation performance without additional training, and many studies have attempted to improve the translation quality and the decoding speed~\cite{zheng-etal-2021-adaptive,meng-etal-2022-fast,wang-etal-2022-efficient,martins-etal-2022-chunk,dai-etal-2023-simple,deguchi-etal-2023-subset}.
\knnmt stores translation examples in a datastore, which is represented by pairs of a key vector and a value token.
However, it is computationally expensive and time-consuming 
in both constructing a datastore and retrieving the \knn examples because the datastore size is the number of target tokens in the parallel data.

\begin{figure}[t]
    \centering
    \includegraphics[width=0.85\linewidth]{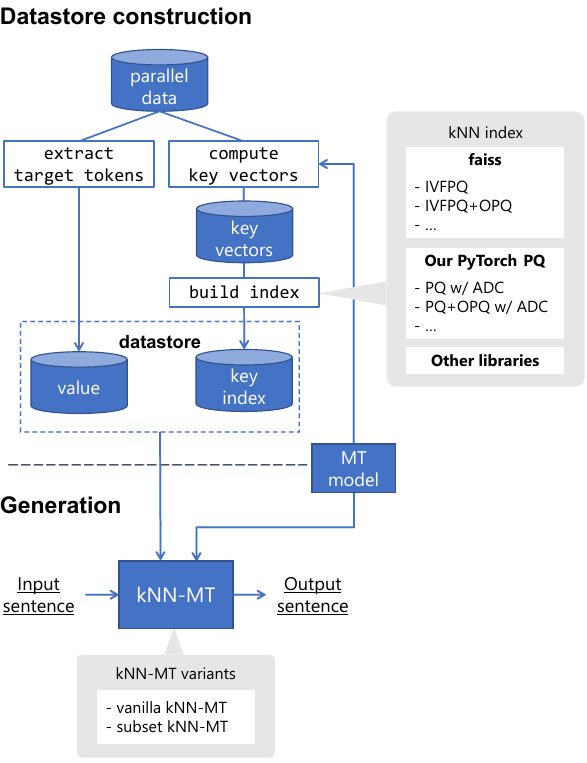}
    \caption{Flow diagram of \knnseq.}
    \label{fig:knnseq}
\end{figure}

This paper presents \knnseq, an efficient and extensible \knnmt framework for researchers and developers.
\knnseq is easy to switch models by developing as a \fairseq plug-in~\cite{ott-etal-2019-fairseq}.
The flow diagram of \knnseq is shown in Figure~\ref{fig:knnseq}.
For datastore construction, we can use arbitrary \knn search libraries, not only \faiss~\cite{johnson-etal-2019-billion}.
For generation, subset \knnmt~\cite{deguchi-etal-2023-subset} can be used as well as vanilla \knnmt.
In addition, we provide \faiss wrapped \knn index which is carefully designed to run efficiently even with a billion-scale large datastore.
Our provided index overrides the internal behavior of \faiss, and it makes significantly faster than the naive implementation.

The experimental results show that our \knnseq constructed a billion-scale datastore in 2.21 hours and achieved comparable gain to the original paper~\cite{khandelwal-etal-2021-nearest} in the WMT'19 German-to-English translation task and the domain adaptation task.

\section{Background}
\subsection{\knnmt}

\paragraph{Datastore construction}

Before decoding, \knnmt constructs a datastore, which stores the translation examples to be accessed during generation.
Let $\bm{x} = (x_1, \ldots, x_{|\bm{x}|}) \in \mathcal{V}_X^{|\bm{x}|}$ and $\bm{y} = (y_1, \ldots, y_{|\bm{y}|}) \in \mathcal{V}_Y^{|\bm{y}|}$ denote a source sentence and target sentence, respectively, where $\bm{|\cdot|}$ is the length of the sequence, and $\mathcal{V}_X$ and $\mathcal{V}_Y$ are the vocabularies of the source language and the target language, respectively.

\knnmt stores translation examples as a set of key--value pairs.
Each target token $y_t$ from the translation examples is stored in the datastore with a $d$ dimensional key ($\in \mathbb{R}^d$), which is the representation of the translation context $(\bm{x}, \bm{y_{<t}})$ obtained from the decoder of the pre-trained NMT model.

The datastore $\mathcal{M} \subseteq \mathbb{R}^d \times \mathcal{V}_Y$ is formally defined as follows:
\begin{equation}
    \mathcal{M} = \{ ( f(\bm{x}, \bm{y}_{<t}), y_t ) \mid (\bm{x}, \bm{y}) \in \mathcal{D}, 1 \leq t \leq |\bm{y}| \},
\end{equation}
where $\mathcal{D}$ is parallel data and $f: \mathcal{V}_X^{|\bm{x}|} \times \mathcal{V}_Y^{t-1} \to \mathbb{R}^d$ is a function that computes the intermediate representation of the final decoder layer from the source sentence and prefix target tokens and employs the representation before passing into to the final feed-forward layer is used as the key vector~\cite{khandelwal-etal-2021-nearest}.

\paragraph{Generation}
During decoding, \knnmt retrieves the \knearestneighbor key--value pairs $\{ (\bm{k}_i, v_i) \}_{i=1}^k \subseteq \mathbb{R}^d \times \mathcal{V}_Y$ from the datastore $\mathcal{M}$ using the query vector $f(\bm{x}, \bm{y}_{<t})$ at timestep $t$.
Next, $\pknn$ is calculated as follows:
\begin{align}
    \label{eq:pknn-num}
    &\pknn(y_t | \bm{x}, \bm{y}_{<t}) \nonumber \\
    &\propto \sum_{i=1}^k \mathbbm{1}_{y_t=v_i} \exp{\frac{-\lVert \bm{k}_i - f(\bm{x}, \bm{y}_{<t}) \rVert^2_2}{\tau}},
\end{align}
where $\tau$ is the temperature parameter for $\pknn$.
Then, \knnmt generates the output probability by computing the linear interpolation between the \knn and MT probabilities, $\pknn$ and $\pmt$:
\begin{align}
    \label{eq:prob-out}
    &P(y_t | \bm{x}, \bm{y}_{<t}) \nonumber \\
    &= \lambda \pknn(y_t | \bm{x}, \bm{y}_{<t}) + (1-\lambda) \pmt(y_t | \bm{x}, \bm{y}_{<t}),
\end{align}

\subsection{IVFPQ for Billion-Scale \knn Search}
The size of the datastore depends on the number of all target tokens in parallel data, often in billion-scale.\footnote{
In our experiment, we construct the datastore from 
997.7M
tokens.
}
The original \knnmt~\cite{khandelwal-etal-2021-nearest} and subsequent studies~\cite{zheng-etal-2021-adaptive,meng-etal-2022-fast,wang-etal-2022-efficient,martins-etal-2022-chunk,deguchi-etal-2023-subset} employ inverted file index with product quantization (IVFPQ)~\cite{jegou-etal-2011-product} for the approximate nearest neighbor (ANN) search to reduce the memory footprint and improve the search speed.
Note that there still remains a problem of search space size, thus having room for speed improvements.

\paragraph{Inverted file index (IVF)}
IVF is a data structure used to improve the search speed.
IVF performs k-means clustering on all vectors and stores a mapping from the centroid vector of each cluster to the vectors belonging to that cluster.
During search, IVF first finds the $n$-nearest-neighbor centroid vectors, and then retrieves the 
\knearestneighbor vectors from the subset of vectors belonging to those clusters.
Note that the computational cost of k-means clustering, which is used for constructing an IVF, is $\mathcal{O}(NI)$ where $N$ is the number of vectors, i.e., $|\mathcal{M}|$ in \knnmt, and $I$ is the number of k-means iterations.
Thus, IVF becomes computationally heavy when $\mathcal{M}$ has billions of vectors.\footnote{
In \faiss, k-means computation of IVF is performed on sampled vectors to reduce computational cost; however, it is still slow.
}

\paragraph{Product Quantization (PQ)}
PQ is a method of vector quantization to reduce the memory footprint.
PQ~\cite{jegou-etal-2011-product} splits a $d$ dimensional vector into $M$ sub-vectors and quantizes each $\frac{d}{M}$ dimensional sub-vector.
Codewords are learned by k-means clustering of key vectors in each subspace.
The codewords of the $m$-th sub-space $\mathcal{C}^{m}$ are formulated as follows:
\begin{equation}
    \mathcal{C}^{m} = \{ \bm{c}_{1}^{m}, \ldots, \bm{c}_{L}^{m}\},~ \bm{c}_{l}^{m} \in \mathbb{R}^{\frac{d}{M}}.
\end{equation}
The typical PQ has $L=256$ codewords for each subspace, and a quantized code is represented by an unsigned 8-bit integer (\texttt{uint8}).
When a vector $\bm{u} \in \mathbb{R}^d$ is quantized, its code vector $\bar{\bm{u}}$ is calculated as follows:
\begin{align}
    \bar{\bm{u}} &= [\bar{u}^{1}, \ldots, \bar{u}^{M}]^{\top} \in \{1, \ldots, L\}^{M},\\
    \bar{u}^{m} &= \argmin_{l} \| \bm{u}^m - \bm{c}_{l}^{m} \|_{2}^{2},~\bm{u}^{m} \in \mathbb{R}^{\frac{d}{M}}.
\end{align}

\paragraph{Inverted file index with product quantization (IVFPQ)}
IVFPQ~\cite{jegou-etal-2011-product} is used to reduce the memory footprint and improve the search speed by combining IVF and PQ.
The quantized vector of IVFPQ is different from that of general PQ in that the residual representation is used between the data and the centroid vector obtained by k-means clustering of IVF.

\subsection{Subset \knnmt}
Subset \knnmt~\cite{deguchi-etal-2023-subset} addresses the problem of decoding speed of billion-scale \knnmt, which reduces the search space of \knnmt by retrieving the neighboring sentences of the input sentence.

\paragraph{Subset retrieval}
Firstly, a sentence datastore $\mathcal{S}$ is constructed as follows:
\begin{align}
    \mathcal{S} &= \{ (s(\bm{x}), \bm{y}) \mid (\bm{x}, \bm{y}) \in \mathcal{D} \},
\end{align}
where $s: \mathcal{V}_{X}^{|\bm{x}|} \to \mathbb{R}^{d'}$ is a sentence encoder, which computes a $d'$ dimensional vector representation of a source sentence.
Before starting the decoding step, the model retrieves the \nnearestneighbor sentences of the input sentence from the sentence datastore $\mathcal{S}$.
Let $\hat{\mathcal{S}} \subset \mathcal{S}$ be the subset comprising \nnearestneighbor sentences.
The search space for target tokens in \knnmt is then drastically reduced by constructing the datastore corresponding to $\hat{\mathcal{S}}$ as follows:
\begin{align}
    \hat{\mathcal{M}} = \{ &(f(\bm{x}, \bm{y}_{<t}), y_{t}) \mid \nonumber \\
    &(s(\bm{x}), \bm{y}) \in \hat{\mathcal{S}}, 1 \leq t \leq |\bm{y}| \},
\end{align}
where $\hat{\mathcal{M}} \subset \mathcal{M}$ is the reduced datastore for the translation examples coming from the \nnearestneighbor sentences.
During decoding, the model uses the same algorithm as \knnmt except that $\hat{\mathcal{M}}$ is used as the datastore instead of $\mathcal{M}$.

\paragraph{Distance Look-Up Table}
In subset \knnmt, the search space varies dynamically depending on the input sentence.
Therefore, an IVF cannot be used; instead, it is necessary to calculate the distance for each key in the subset.
For this purpose, we use asymmetric distance computation (ADC)~\cite{jegou-etal-2011-product} instead of the usual distance computation between floating-point vectors.
In ADC, the number of table lookup is linearly proportional to the number of keys $N$ in the subset.
ADC is an efficient method for computing the distance between a query vector $\bm{q} \in \mathbb{R}^d$ and $N$ key codes
$\bar{\mathcal{K}} = \{ \bar{\bm{k}}_i \}_{i=1}^{N} \subseteq \{ 1, \ldots, L \}^M$.
The distance look-up table (LUT) $\bm{A}^m \in \mathbb{R}^{L}$ is computed by calculating the distance between a query $\bm{q}^m$ and the codes $\bm{c}_l^m \in \mathcal{C}^m$ in each subspace $m$, as follows:
\begin{equation}
    \label{eq:adc-dt}
    A^m_l = \| \bm{q}^m - \bm{c}_{l}^{m} \|_{2}^{2}.
\end{equation}
Then, the distance between a query and each quantized key $d(\bm{q}, \bar{\bm{k}}_i)$ is obtained by looking up the distance LUT as follows:
\begin{equation}
    \label{eq:adc-lu}
    d(\bm{q}, \bar{\bm{k}}_i) = \sum_{m=1}^{M} d_m(\bm{q}^m, \bar{k}_i^m) = \sum_{m=1}^M A^m_{\bar{k}_i^m}.
\end{equation}
A LUT in each subspace, $\bm{A}^m \in \mathbb{R}^L$, consists of the distance between a query and codes.
The number of codes in each subspace is $L$, and the distance is a scalar; therefore, $\bm{A}^m$ has $L$ distances.
The look-up key is the code ID of a quantized key itself, i.e., if the $m$-th subspace's code of a key is $5$, ADC looks-up $A^m_{5}$.
By using ADC, the distance is computed only once\footnote{
The direct distance computation requires $N$ times calculations according to $\lVert \bm{q} - \bm{k}\rVert^2$.
ADC computes the distance only $L \ll N$ times and just looks-up the LUT $N$ times.
}~(Equation \ref{eq:adc-dt}) and does not decode PQ codes into $d$ dimensional key vectors;
therefore, it can compute the distance while keeping the key in the quantization code, then the \knearestneighbor tokens are efficiently retrieved from $\hat{\mathcal{M}}$.

\section{Our Framework: \knnseq}

\textsc{knn-seq} is designed to be extensible in terms of models, methods, and \knn indexes, and also designed to run efficiently.
Each of the implemented components, such as models, indexes, and utilities, has been confirmed to pass the unit tests.

\subsection{Extensibility}
\knnseq is designed to be extensible: it can switch easily between different models and \knn indexes.

\paragraph{Model}
Our framework is built on top of \fairseq~\cite{ott-etal-2019-fairseq} as a plug-in rather than a fork.
This enables our implementation to work seamlessly with \fairseq while being unaffected by upstream changes.
The main \knnmt computations, i.e., Equation~\ref{eq:pknn-num} and \ref{eq:prob-out}, are implemented in \verb|EnsembleModel| which enables an ensemble of  any kind of encoder-decoder or decoder-only models, allowing for customized models and ensemble decoding.

\paragraph{\knn index}
\knnseq was designed for flexibility and ease of switching the \knn search index.
The \verb|SearchIndex| abstract class enables various \knn search libraries and search strategies to be implemented simply by inheriting the class and wrapping the necessary methods.
The index used for the datastore can be switched easily.
As a default choice of the \knn index, we wrapped \faiss~\cite{johnson-etal-2019-billion}.

\faiss and other almost all \knn search libraries only support the full set search and cannot search from the subset dynamically created online~\cite{matsui-etal-2018-reconfigurable};
therefore, we implement a PyTorch-based subset search index for subset \knnmt.
\knnseq can be easily switched to this index.

\subsection{Efficiency}
\paragraph{Datastore construction}
The datastore is constructed in three steps: 
(1) store the value tokens, (2) compute the key vectors using a trained NMT model, (3) build the \knn index for efficient search.
\knnseq makes the two most time consuming steps, key vector computation and datastore indexing, more efficient.

The value tokens are stored by Hierarchical Data Format version 5 (HDF5)\footnote{\url{http://www.hdfgroup.org/HDF5}}, designed to store and organize large amounts of data.
Key vectors are computed by feeding the stored value tokens and their source tokens into a trained NMT model.
In our implementation, the value tokens are ordered by their sequence lengths; thus, it reduces the number of paddings in a mini-batch and accelerates the key vector computation.
The key vectors are also stored in HDF5.

Then, the \knn index is built from key vectors.
We implemented a \faiss wrapped index as the default index.
Our design allows for the construction of indexes of various sizes, from small to large, but especially it is designed to ensure that billion-scale indexes can be constructed efficiently.
Our \faiss wrapped index overrides the internal behavior of \faiss to allow for several time-consuming processes to be run on the GPU, including the IVF k-means clustering, the PQ codewords learning, and vector addition into the index.
This significantly increases the speed of building an index on GPU.

\paragraph{Vector pre-transformation}
\knnseq makes dimension reduction with PCA (Principal Component Analysis) or pre-transformation with OPQ (Optimized PQ)~\cite{ge-etal-2014-optimized} more efficient as well.
These methods are often applied to reduce the computational cost or improve the search accuracy of a datastore.

Dimension reduction with PCA $\mathrm{PCA}: \mathbb{R}^d_\mathrm{in} \to \mathbb{R}^{d_\mathrm{out}}$ is computed as follows:
\begin{equation}
    \mathrm{PCA}(\bm{u}) = \bm{W}^\mathrm{PCA}(\bm{u} - \bm{\mu}),
\end{equation}
where $\bm{W}^\mathrm{PCA} \in \mathbb{R}^{d_\mathrm{out} \times d_\mathrm{out}}$ is the component matrix learned from data and $\bm{\mu} \in \mathbb{R}^d$ is the averaged values for each row computed from the data vectors.
Pre-transformation with OPQ $\mathrm{OPQ}: \mathbb{R}^d \to \mathbb{R}^d$ is computed as follows:
\begin{equation}
    \mathrm{OPQ}(\bm{u}) = \bm{W}^\mathrm{OPQ} \bm{u},
\end{equation}
where $\bm{W}^\mathrm{OPQ} \in \mathbb{R}^{d \times d}$ is the rotation matrix which is learned to minimize the quantization error of PQ.

We implement their computation using PyTorch to enable GPU acceleration, as both dimension reduction with PCA and pre-transformation with OPQ can be represented as linear projections and elementwise subtraction.

In addition, we implemented OPQ training to take advantage of the GPU.
The PQ codewords are iteratively trained during training OPQ, which is time-consuming; thus, \knnseq uses a GPU in this step by overriding the internal behavior of \faiss.
The rotation matrix of OPQ $\bm{W}^\mathrm{OPQ}$ is trained by computing these procedures iteratively:
(1) training the PQ codewords, (2) calculating the reconstruction error using the trained PQ codewords, and (3) updating the rotation matrix to minimize the reconstruction error.

\paragraph{Generation}
The decoding speed of \knnmt is time-consuming and two orders of magnitude slower than the base MT model~\cite{khandelwal-etal-2021-nearest}.
Our \knnseq can transfer the billion-scale IVFPQ to multiple GPUs by distributing shard indexes to speed up generation.

\section{Experiments}
We conducted translation experiments using our \knnseq and evaluated the translation performance and efficiency.
We measured the translation performance by BLEU, chrF, and COMET, and the decoding speed by the number of tokens generated per second (tok/s).
We used \textsc{sacreBLEU} to calculate BLEU
and chrF.
We used a pre-trained NMT model provided by \fairseq as the base MT.
The MT model is Transformer big with $d=1024$ dimensional embeddings~\cite{ng-etal-2019-facebook}.
All models generated translations using beam search with a beam size of 5 and a length penalty of $1.0$.
In \knnmt, the 32 nearest neighbor clusters were searched by the IVF, and $k=64$ nearest neighbor tokens were retrieved.
The datastores were constructed using 8 NVIDIA V100 GPUs and 16 core CPUs.\footnote{Intel(R) Xeon(R) Gold 6150 CPU @ 2.70GHz}
We varied GPU resource settings and batch sizes while measuring the decoding speed:
either using 8 V100 GPUs (G$^8_*$) or a single V100 GPU (G$^1_*$) for GPU resources, and 12,000 tokens ($*_{\infty}$) or a single sentence ($*_1$) for batch sizes.
Because the \knn index does not fit into a single GPU memory due to the large index size, \knn search of \knnmt with GPU use 8 GPUs, i.e., G$^8_*$, and other settings use only a single GPU.

\subsection{WMT'19 German-to-English Translation}
\label{sec:exp-wmt19}

We evaluated our \knnmt on the WMT'19 German-to-English (De-En) translation task using a billion-scale datastore.
The datastore was constructed from the parallel data of the WMT'19 De-En news translation task with subword lengths of 250 or less and a sentence length ratio of 1.5 or less between the source and target sentences.
The datastore contained 
997.7M
target tokens obtained from 
37.0M
sentence pairs.
We employed IVFPQ for the \knn search.
The number of IVF centroids was set to 131,072 and the number of PQ subspaces was set to $M=64$.
We experimented with constructing the index using OPQ vector pre-transformation.
In subset \knnmt, the key vectors of the sentence datastore are computed by averaging the hidden vectors from the encoder of the MT model, and IVFPQ was used for the index with 32,768 IVF centroids, and $M=64$ PQ subspaces, using OPQ.
The target token datastore of subset \knnmt was quantized by PQ with $M=64$ subspaces using OPQ.
Subset \knnmt retrieved 512 nearest neighbor source sentences for each input sentence.
In both \knnmt and subset \knnmt, the temperature parameter $\tau$ was set to 100, and the weight of $\pknn$, $\lambda$, was set to 0.4.
We evaluated translation performance on newstest2019.

The processing time of datastore construction is shown in Table~\ref{tab:build-ds}.
As shown in the table, the billion-scale datastore construction is completed in 2.21 hours by using our \knnseq.
The datastore construction time is only increased by 30 minutes even if using OPQ in \knnseq.

Table~\ref{tab:result-wmt19deen} shows the translation performance and the decoding speed.
The \knnmt improved $+1.4\%$ BLEU, $+0.8\%$ chrF, and $+0.1\%$ COMET scores over the base MT.
Our implementation achieved gains comparable to the original paper~\cite{khandelwal-etal-2021-nearest}.
The results also show that using multiple GPUs for the \knn search during inference is faster than using the CPU index.
When using OPQ, the translation performance improved a bit further.
In this experiment, subset \knnmt achieved $+1.0\%$ BLEU, $+0.3\%$ chrF, and $+0.1\%$ COMET scores compared to the base MT.
And also, we confirmed that subset \knnmt runs on a single GPU and the decoding was finished in roughly 92\% and 49\% speeds of the base MT in G$^1_1$ and G$^1_\infty$ settings, respectively.

\begin{table}[t]
    \centering
    \small
    \begin{tabular}{@{}lrr@{}} \toprule
         & \multicolumn{2}{c}{Time (sec)} \\ \cmidrule(l){2-3}
         & w/o OPQ & w/ OPQ \\ \midrule
        Compute keys & 1539.3 & 1539.3 \\
        Train IVFPQ & 326.0 & 843.7 \\
        Build IVFPQ & 6104.0 & 7394.5 \\ \midrule
        Total & 7969.3 & 9777.5 \\
         & (2.21 h) & (2.72 h) \\\bottomrule
    \end{tabular}
    \caption{Processing time of billion-scale datastore construction for \knnmt in the WMT'19 De-En translation task.
    }
    \label{tab:build-ds}
\end{table}
\begin{table*}[t]
    \centering
    \small
    \begin{tabular}{@{}lrrrrrrr@{}} \toprule
         & \multicolumn{3}{c}{Quality} & \multicolumn{4}{c}{Speed ($\uparrow$tok/s)} \\ \cmidrule(lr){2-4} \cmidrule(l){5-8}
         & BLEU & chrF & COMET & G$^1_1$ & G$^1_\infty$ & G$^8_1$ & G$^8_\infty$ \\ \midrule
        Base MT & 39.5 & 64.0 & 84.6 & 136.4 & 3133.6 & --- & --- \\
        \knnmt & 40.9 & 64.8 & 84.7 & 1.4 & 4.7 & 75.0 & 555.1 \\
        ~~~+ OPQ & 41.1 & 65.0 & 84.9 & 1.1 & 4.8 & 67.9 & 518.4 \\
        Subset \knnmt & 40.5 & 64.3 & 84.7 & 126.0 & 1539.2 & --- & ---  \\
        \bottomrule
    \end{tabular}
    \caption{Results of the WMT'19 De-En translation task.}
    \label{tab:result-wmt19deen}
\end{table*}

\subsection{Domain Adaptation}
\begin{table*}[t]
    \centering
    \small
    \begin{tabular}{@{}lrrrrrrrr@{}} \toprule
    & \multicolumn{2}{c}{IT (3.1M)} & \multicolumn{2}{c}{Koran (449.6K)} & \multicolumn{2}{c}{Law (18.3M)} & \multicolumn{2}{c}{Medical (5.7M)} \\
    & \multicolumn{2}{c}{$\tau=10, \lambda=0.7$} & \multicolumn{2}{c}{$\tau=100, \lambda=0.8$} & \multicolumn{2}{c}{$\tau=10, \lambda=0.8$} & \multicolumn{2}{c}{$\tau=10, \lambda=0.8$}  \\
    \cmidrule(lr){2-3} \cmidrule(lr){4-5} \cmidrule(lr){6-7} \cmidrule(l){8-9} 
         & BLEU & tok/s & BLEU & tok/s & BLEU & tok/s & BLEU & tok/s \\ \midrule
        Base MT & 37.9 & 2819.4 & 16.9 & 3042.8 & 45.9 & 2831.7 & 40.3 & 2813.3 \\
        CPU-\knn \knnmt & 45.7 & 434.1 & 21.1 & 730.4 & 62.0 & 151.4 & 55.2 & 341.6 \\
        GPU-\knn \knnmt & 45.7 & 2229.3 & 21.1 & 2567.4 & 62.0 & 1848.1 & 55.2 & 2079.1 \\
        \bottomrule
    \end{tabular}

    \caption{Results of the De-En domain adaptation task. The chrF and COMET scores are listed in the appendix. The right of the domain name shows each datastore size, i.e., the number of target tokens in each in-domain data.}
    \label{tab:result-domain-deen}
\end{table*}

We also evaluated the out-of-domain translation tasks in the IT, Koran, Law, and Medical domains~\cite{koehn-knowles-2017-six,aharoni-goldberg-2020-unsupervised}.
The datastores of each domain were constructed from in-domain parallel data.
The MT model is the same as the one used in Section~\ref{sec:exp-wmt19}.
The decoding speed is measured in the G$^1_\infty$ setting.
In this experiment, we also compared the decoding speed between using the GPU index and the CPU index in \knnmt because each datastore has only 100K order sentences, and the index can be transferred to a single V100 GPU.

Table~\ref{tab:result-domain-deen} shows the results of the domain adaptation task.
The results show that our \knnmt implementation achieved almost comparable scores to the original paper~\cite{khandelwal-etal-2021-nearest}.
The results also show that it can work on a single GPU when the datastore is small, and using a GPU for \knn search improves the decoding speed by 3 to 10 times.

\section{Related Work}
\knnmt has also been used for grammatical error correction to improve its accuracy or interpretability~\cite{kaneko-etal-2022-interpretability, vasselli-watanabe-2023-closer}.
\knnseq can be employed in any encoder-decoder models and it is easy to use for other generation tasks than machine translation.

There are several existing frameworks for \knnmt, including \knn-BOX~\citep{zhu-etal-2023-knnbox} and knn-transformers~\citep{alon-etal-2022-neuro}.\footnote{
\url{https://github.com/neulab/knn-transformers}}
Despite the advancements in \knn-based translation methods, many prior works~\cite{zheng-etal-2021-adaptive,meng-etal-2022-fast,martins-etal-2022-chunk,dai-etal-2023-simple} have utilized the original implementation~\cite{khandelwal-etal-2021-nearest} or adaptive \knn models~\cite{zheng-etal-2021-adaptive}, both of which fork \fairseq, making them more challenging to maintain and update. 
In contrast, our implementation is a novel extension of \fairseq that does not require forking, resulting in a more flexible and maintainable framework. 
Moreover, our framework addresses the computational challenges by offering options for more efficient indexing, making it a viable solution for real-world applications and scalable to large datastores.

\knnseq can employ other \knn search libraries and algorithms~\cite{muja-and-lowe-2009-fast,boytsov-etal-2013-engineering,malkov-and-yashunin-2020-efficient}.

\section{Conclusion}
We presented an efficient and extensible \knnmt framework, \knnseq, for researchers and developers.
\knnseq is developed as a \fairseq plug-in and was carefully designed to make it easy to switch models and \knn indexes.
In addition, our \faiss wrapped \knn index is designed to run efficiently even with a billion-scale large datastore.
Experimental results show that our \knnmt achieved a comparable gain to the original paper and the billion-scale datastore construction took only 2.21 hours in the WMT'19 German-to-English translation task.
In future work, we would like to try to use other \knn search algorithms.
This paper experimented with IVFPQ and IVFPQ+OPQ, but further \knn search methods can be used to improve the speed or accuracy of IVFPQ.
Hierarchical navigable small world (HNSW)~\cite{malkov-and-yashunin-2020-efficient} is a graph-based search algorithm that can search fast and accurately on CPUs, which is useful for million-scale \knn search like small-scale datastore.
It can also be combined with IVFPQ by using HNSW in the coarse search of IVF, which achieved the state-of-the-art performance of \knn search.
We hope that \knnseq will accelerate the experimental cycle of studies using \knnmt.

\section*{Limitations}
In \knnseq, the computational complexity has not been reduced.
While we optimized the implementation, we did not attempt to improve the method.
Some \knnseq accelerators require GPU resources.
Training the IVF of an IVFPQ and transferring shard indexes may require multiple GPUs.
The speed gain may depend on the GPU model number and hardware configuration.

\section*{Ethics Statement}
If the parallel data for constructing the datastore contains toxic text, \knnmt has the risk of generating toxic content.

\bibliography{anthology,custom}
\bibliographystyle{acl_natbib}

\appendix

\section{Datasets, Tools, Models}

\paragraph{Datasets}
Parallel data of the WMT'19 De-En translation task can be used for research purposes as described in \url{https://www.statmt.org/wmt19/translation-task.html}.
The five domain adaptation datasets in De-En can be used for research purposes as described in the paper~\cite{aharoni-goldberg-2020-unsupervised}.

\paragraph{Tools}
\fairseq and \faiss are MIT-licensed.

\paragraph{Models}
We used \texttt{model1.pt} that is included in \url{https://dl.fbaipublicfiles.com/fairseq/models/wmt19.de-en.joined-dict.ensemble.tar.gz} for the De-En MT model which is included in \fairseq and it is MIT-licensed.
To evaluate COMET scores, we used \texttt{Unbabel/wmt22-comet-da}.

\section{Details of Translation Quality}
The chrF and COMET scores of the De-En domain adaptation task are shown in Table~\ref{tab:result-domain-deen-extra}.

\begin{table*}[t]
    \centering
    \small
    \begin{tabular}{@{}lrrrrrrrr@{}} \toprule
    & \multicolumn{2}{c}{IT} & \multicolumn{2}{c}{Koran} & \multicolumn{2}{c}{Law} & \multicolumn{2}{c}{Medical} \\
    \cmidrule(lr){2-3} \cmidrule(lr){4-5} \cmidrule(lr){6-7} \cmidrule(lr){8-9}
         & chrF & COMET & chrF & COMET & chrF & COMET & chrF & COMET  \\ \midrule
        Base MT & 58.5 & 82.2 & 39.8 & 72.3 & 66.1 & 85.4 & 61.0 & 83.2 \\
        CPU-\knn \knnmt & 62.7 & 83.1 & 42.4 & 73.1 & 76.3 & 87.0 & 70.0 & 84.6 \\
        GPU-\knn \knnmt & 62.8 & 83.2 & 42.5 & 73.1 & 76.2 & 87.0 & 70.0 & 84.6 \\
        \bottomrule
    \end{tabular}
    \caption{The chrF and COMET scores of the De-En domain adaptation task using \knnseq.}
    \label{tab:result-domain-deen-extra}
\end{table*}

\end{document}